\documentclass{article}

\usepackage[preprint]{neurips_2024}

\usepackage[utf8]{inputenc} % allow utf-8 input
\usepackage[T1]{fontenc}    % use 8-bit T1 fonts
\usepackage{hyperref}       % hyperlinks
\usepackage{url}            % simple URL typesetting
\usepackage{booktabs}       % professional-quality tables
\usepackage{amsfonts}       % blackboard math symbols
\usepackage{nicefrac}       % compact symbols for 1/2, etc.
\usepackage{microtype}      % microtypography
\usepackage{xcolor}         % colors
\usepackage{subcaption}
\usepackage{graphicx}
\usepackage{amsmath}
\usepackage{bm}
\usepackage{multicol}
\usepackage{float}

\title{Active perception and disentangled representations allow continual, episodic zero and few-shot learning}

\author{%
  David Rawlinson \\
  Cerenaut AI\\
  Melbourne\\
  Australia \\
  \texttt{dave@cerenaut.ai} \\
    \And
   Gideon Kowadlo \\
  Cerenaut AI\\
  Melbourne\\
  Australia \\
   \texttt{gideon@cerenaut.ai} \\
}

\begin{document}

\maketitle

\begin{abstract}
Generalization is often regarded as an essential property of machine learning systems. However, perhaps not every component of a system needs to generalize. Training models for generalization typically produces entangled representations at the boundaries of entities or classes, which can lead to destructive interference when rapid, high-magnitude updates are required for continual or few-shot learning. Techniques for fast learning with non-interfering representations exist, but they generally fail to generalize. Here, we describe a Complementary Learning System (CLS) in which the fast learner entirely foregoes generalization in exchange for continual zero-shot and few-shot learning. Unlike most CLS approaches, which use episodic memory primarily for replay and consolidation, our fast, disentangled learner operates as a parallel reasoning system. The fast learner can overcome observation variability and uncertainty by leveraging a conventional slow, statistical learner within an active perception system: A contextual bias provided by the fast learner induces the slow learner to encode novel stimuli in familiar, generalized terms, enabling zero-shot and few-shot learning. 
This architecture demonstrates that fast, context-driven reasoning can coexist with slow, structured generalization, providing a pathway for robust continual learning.
\end{abstract}

\begin{figure}[htbp]
     \centering
     % First Row
     \begin{subfigure}[b]{0.32\linewidth}
         \centering
         \includegraphics[width=\linewidth]{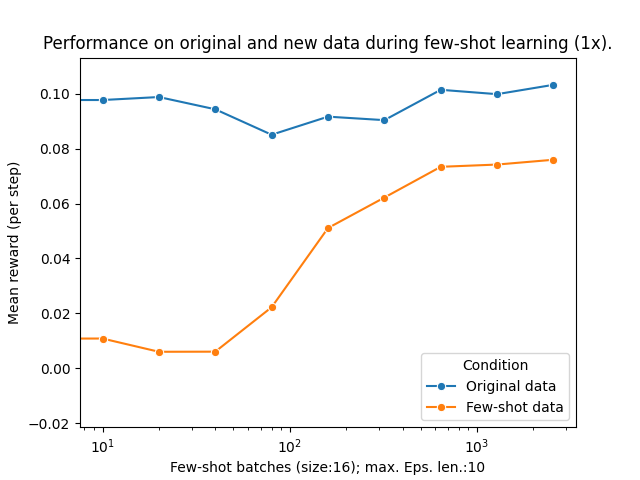}
         \caption{The proposed system can \textbf{few-shot} learn new classes (orange) without interference with existing memories (blue).}
         \label{fig:intro_few_shot}
     \end{subfigure}
     \hfill
     \begin{subfigure}[b]{0.32\linewidth}
         \centering
         \includegraphics[width=\linewidth]{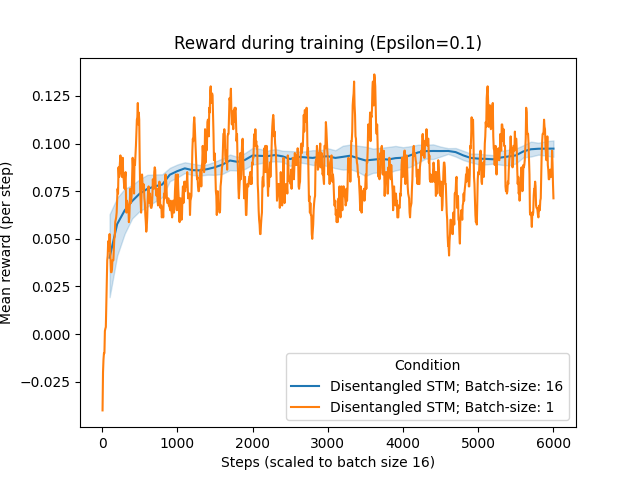}
         \caption{The model can learn in an \textbf{online, streaming} setting with comparable sample efficiency to mini-batches, without replay.}
         \label{fig:streaming_learning}
     \end{subfigure}
     \hfill
     \begin{subfigure}[b]{0.32\linewidth}
         \centering
         \includegraphics[width=\linewidth]{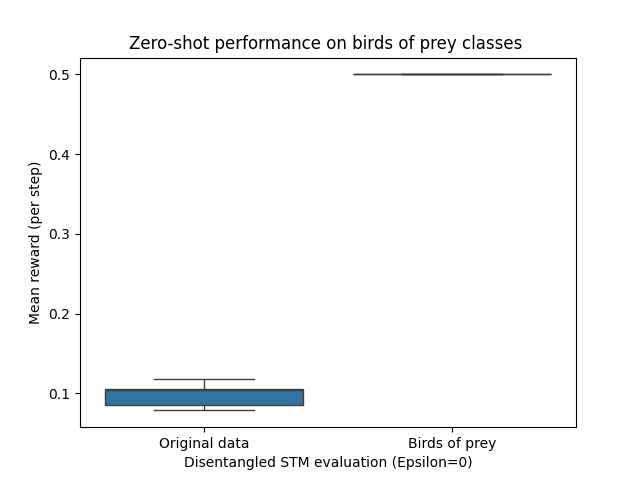}
         \caption{The system can \textbf{zero-shot} new classes if it has already learned similar classes which share observable features and behavioural statistics.}
         \label{fig:intro_zero_shot}
     \end{subfigure}
     
     \caption{A disentangled, fast-learning episodic RL model which entirely delegates generalization to a slow, statistical long-term memory (LTM) avoids the limitations of conventional statistical learning. Further explanation of these results are provided in section \ref{sec:results}.}  
     \label{fig:intro}
\end{figure}

\section{Introduction}
\subsection{Continual, lifelong, zero and few-shot learning}
Continual and lifelong learning refer to the idea that a machine learning model should not just be pre-trained once and then used for inference. Instead, it should continue to be updated with new data as it becomes available, adapting to the changing statistics of the new data including novel, unseen classes and features \citep{parisi2019continual}.

%\subsubsection{Few-shot learning and forgetting}
Few-shot learning is the ability to learn from just a few exposures to new data. The challenge becomes even more difficult in \emph{continual} few-shot learning due to the ongoing introduction of new classes or changing statistics. There are also many variations to this challenge, including when and how to introduce or remove classes from the training data \citep{antoniou2020defining}. 

In any learned statistical representation, individual samples are not memorized independently but contribute incrementally to latent variables which jointly represent the data. By capturing statistical regularities in the data, the learned representation describes not just observed samples but properties of the population from which they were drawn. Attempts to make substantial changes to the representation in response to a few recent samples also disrupts the representation of older samples, causing a phenomenon known as ``catastrophic forgetting'' \citep{goodfellow2013empirical}. Researchers have invented many techniques to minimize this tension, but have not yet overcome it \citep{kirkpatrick2017overcoming}.

%\subsubsection{Zero-shot and meta-learning}
Zero-shot learning - learning without exposure to \emph{any} exposures to new data - sounds impossible at first, but there are two primary approaches: Meta-learning \citep{schmidhuber1987evolutionary, finn2017model} and world-models. In meta-learning, the models learn a constant process of adaptation to current data. ``In-context-learning'' in Large Language Models is a form of meta-learning \citep{brown2020language}. World models allow zero-shot learning by allowing exploration within a learned simulation of the problem. Therefore, it is possible to discover the implications of new data within a world model before encountering it in actual data \citep{zhou2024dino}.

\subsection{Abandoning generalization}
This paper proposes a system in which the primary actor embraces a total lack of generalization ability. In doing so, it gains the ability to learn rapidly without interference between memories. 

Since in all real environments there are various sources of noise, variance and uncertainty, the system must still generalize its inputs. This is achieved using an active perception framework \citep{bajcsy2018revisiting} in which the primary actor questions a slow, statistical learner about current input and context. With its two memories the framework is an instance of a Complementary Learning System (CLS) \citep{kumaran2016learning}, but unlike most prior instantiations \cite{spens2024generative,sun2023organizing,schapiro2017complementary}, the fast memory does not augment the slow one, or help it to learn. Instead, the slow memory supports the fast memory, which is the primary actor.

The system is framed as an episodic Reinforcement Learning (RL) agent which seeks to maximize its environmental reward under changing conditions. This setting is known as continual RL \citep{khetarpal2022towards}. We demonstrate that under these conditions the agent is capable of fast, continual, zero-shot and few-shot learning without forgetting. To achieve this, the agent learns to ask for sequences of perceptions which are task-relevant and invariant to observational variance. The remainder of the introduction provides context and detail about these concepts.

\subsection{Generalization and entanglement}
Machine learning research has long emphasized the importance of generalization: the ability to produce correct predictions for unseen inputs. However, this objective often comes at a cost. Training systems to generalize typically produces entangled representations ~\citep{higgins2017beta}, in which multiple features or entities are overlapping in the latent space. Representations can also be fractured, meaning that key concepts are distributed among many latent variables \citep{kumar2025questioning}. Entanglement and fracturing creates interference and fragility, so that updating the model for one experience can unintentionally disrupt representations of previously learned experiences. This makes continual, few-shot learning difficult \citep{mcclelland1995why,french1999catastrophic,goodfellow2013empirical}. This tension between the desire for generalization and the need for stability is a core challenge for fast learning in neural networks.

\subsubsection{Orthogonality and non-interference in sparse representations}
%Disentangled representations, by contrast, reduce interference and make rapid learning tractable. At the extreme, a hashmap-like memory provides a fully disentangled storage mechanism: each key maps independently to a stored value. However, without compression and integration data, the representation has limited utility.

Sparse representations provide a principled approach to achieving more efficient and robust non-interference between memories. By encoding information such that only a small subset of units is active for a given observation, sparse codes naturally separate features and minimize overlap between representations - they are highly orthogonal. Sparse representations enable fast learning because updates to one subset of active units minimally affect others, a property first formalized in Kanerva's sparse distributed memory~\cite{kanerva1988sparse}. More recently, the Kanerva Machine~\cite{wu2018kanerva} demonstrated that sparse, distributed memory can support generative reconstruction and fast adaptation to new inputs. These findings substantiate the key point that non-interference via sparsity is a powerful enabler of rapid learning, particularly in situations where generalization is not strictly required.

Sparse, distributed representations also have combinatoric capacity, enabling them to represent a very large number of patterns without significant interference~\cite{kanerva1988sparse}. 

\subsubsection{Disentangled sparse representations}
But orthogonality and the resulting non-interference is not sufficient to make a representation disentangled. 
 In a disentangled memory, concepts are separated from each other, but samples share a joint representation in terms of learned concepts. For example, an object might be described in terms of colour and size, and these attributes would be reused for different objects.

\citep{lampert2009learning} describe a system that learns to predict unseen object classes from the \emph{attributes} of known classes, without any exposure to the unseen classes - an example of zero-shot learning. For this to work, the unseen classes must share attributes with known classes. This concept can be combined with sparse coding to create non-interfering representations of object attribute combinations - a disentangled representation.

%Hashmap memories scale effectively, with O(1) retrieval time and linear scaling with respect to capacity. Other methods for fast, non-generalizing learning include simple key-value memory networks or episodic memory systems, which prioritize memorization over abstraction, explicitly contrasting with generalization-focused models. These approaches demonstrate that by decoupling features and entities, systems can rapidly store and retrieve new information with minimal interference. In the Differentiable Neural Computer, \citet{graves2016hybrid} demonstrated that structured memory mechanisms can scale to complex data storage and retrieval tasks when paired with neural controllers. 

%\subsubsection{Orthogonality in sparse representations}

\subsection{Complementary Learning Systems}
The Complementary Learning Systems (CLS) framework provides a biologically inspired solution to balancing rapid learning and long-term generalization. It comprises two memory systems, a Short-Term Memory (STM) and a Long-Term Memory (LTM).

In the canonical biologically-inspired formulation, the hippocampus acts as a fast, episodic STM, while the neocortex as LTM gradually accumulates statistical regularities to support slow, stable learning~\citep{mcclelland1995why}.

%\subsubsection{Episodic memorization for replay \& consolidation}
More recent work has emphasized how the hippocampus can support rapid inference over episodic experiences, enabling learning without catastrophic forgetting \citep{kumaran2016learning}. Most CLS models focus on episodic STM memorization for replay and consolidation to the LTM, allowing the slow system to learn generalized structure \citep{spens2024generative}. Others use the STM more directly. \citep{kaiser2017learning} augmented deep neural models with fast key-value memorization for one-shot learning, with latent activations as keys and supervised targets as values. \citep{santoro2016one} likewise describe one-shot learning of character classification on the Omniglot dataset, from just a few examples of each character-class within each alphabet. They used a ``Neural Turing Machine'' - a fast, fully differentiable memory - to augment their neural network model by providing context for classification. The resulting system surpassed human-level performance.

%\subsubsection{Reasoning in novel conditions}
However, the STM could also serve as a parallel reasoning system, enabling rapid adaptation to new observations or tasks without waiting for slow consolidation. The Tolman--Eichenbaum Machine~\citep{whittington2020tolman} features a hippocampal-inspired module that encodes relational structure in episodes, supporting inference and reasoning directly on new experiences, rather than solely functioning as a training signal for the LTM.  

This paper proposes a reversal of priority between STM and LTM, with the STM being the primary actor supported by the LTM. The concept of the STM as an episodic controller has been previously explored by \citep{blundell2016model} and \citep{pritzel2017neural}. Both works place a hippocampal-inspired model in a RL setting, supporting Q-Learning \citep{watkins1992qlearning} models by rapidly learning the returns of specific state, action combinations. Both were evaluated in the Arcade Learning Environment \citep{bellemare13arcade}. 

In the research described above the STM either memorizes rare and new data to support LTM consolidation or is an intrinsic part of an episodic controller capable of fast, context-specific reasoning in rapidly changing conditions. In the former case generalization occurs after consolidation. In \citep{blundell2016model} limited generalization and memory capacity is noted as a weakness of the STM. In \citep{pritzel2017neural} the STM is part of a model that has learned a generalized memory control policy, which might be considered a form of meta-learning. 

In our work, the STM is the primary actor and does not attempt to generalize. Instead, it must somehow leverage the generalization capability of the slow, statistical memory to overcome observation variability, noise and uncertainty. 

\subsection{Active Perception}
Perception is the process by which an intelligent system interprets high-dimensional sensory data to produce more abstract and compressed internal representations of the environment that are useful for decision making and reasoning. 
%In biological and artificial agents alike, perception transforms high‑dimensional raw input into lower‑dimensional structures that are useful for decision making and reasoning. 
%Originally, this was modelled as a bottom‑up flow of information from sensory receptors to higher‑level representations, driven solely by data in the environment. However, empirical and theoretical work in psychology and neuroscience emphasizes that perception is also shaped by top‑down influences, such as expectations, prior knowledge, goals, and contextual cues, which modulate how sensory information is interpreted and which features are extracted from it. ?? 
%(cite https://academic.oup.com/edited-volume/38606/chapter-abstract/334710934?redirectedFrom=fulltext&login=false)

Active perception describes a process in which the agent takes actions to influence or improve its own sensory input. These actions might include moving to a new viewpoint, and changing the way sensor data is interpreted and integrated over multiple observations to resolve ambiguity. 

%Early work in robotics and vision framed active perception as a tight coupling between sensing and action, where the agent’s behavior is selected not only to satisfy a task but to maximize information gain. More formal definitions characterize an active perceiver as one that ``knows why it wishes to sense, and then chooses what to perceive, and determines how, when and where to achieve that perception''. ? 
% https://pmc.ncbi.nlm.nih.gov/articles/PMC6954017/?utm_source=chatgpt.com

%The interplay between bottom‑up and top‑down processes reflects how 
Perception can be both data‑driven and goal‑guided. Bottom‑up perceptual processes build structure from sensory input, whereas top‑down adapts that structure to satisfy task demands, effectively filtering noise and highlighting contextually relevant aspects of the input. Active perception links perception and action in a closed loop: the agent’s current beliefs and goals guide where and how it gathers data, and the resulting data updates its beliefs and subsequent actions. This perspective aligns closely with notions from cognitive neuroscience that perception is not a passive ``snapshot'' of the world but a dynamically constructed interpretation influenced by prior knowledge, expectations, and behavioral goals \citep{bajcsy2018revisiting}.

\subsubsection{Active Perception in a Complementary Learning System}
In a complementary learning system, a disentangled, non-generalizing STM could overcome data variance by using active perception to ensure its input is in generalized, familiar terms, even if the external input is novel.

\begin{figure}[htbp]
     \centering
     \includegraphics[width=0.4\linewidth]{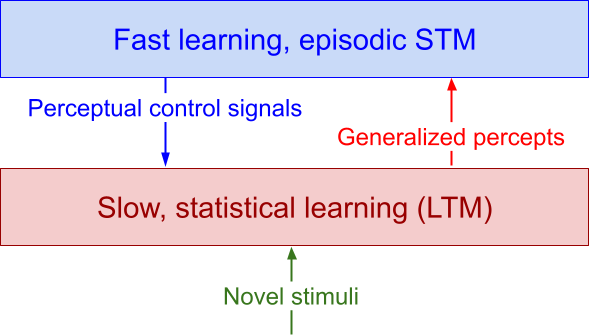}
     \caption{Active perception within a Complementary Learning System allows the STM to learn rapidly without needing to generalize. Generalization occurs within the LTM, which is manipulated by the STM to emit stable, familiar representations relevant to the task. Using the generalization ability within the LTM, the STM can even immediately reason about novel stimuli if they are expressed in terms the STM can already understand.}
     \label{fig:active_perception_with_cls}
\end{figure}

In this model, the STM emits perception control signals which modify the way the LTM represents external input (figure \ref{fig:active_perception_with_cls}). Since the LTM is a slow, statistical learner, it can generalize sensory data into terms the slow system already understands. For instance, the same image of a bird could be perceived at different levels of specificity — “a flying animal,” “a bird,” or “a cockatoo” - or even in task‑relevant terms such as “is this a threat?” These observable features are reminiscent of the object attributes which \citep{lampert2009learning} used to achieve zero-shot learning.

If LTM can return stable, familiar representations invariant to noise and novel features, a fast-learning STM can act immediately (zero-shot) on out-of-domain data without itself needing to generalize. If a change in behaviour is needed, the STM can adapt rapidly, including one-shot.

The STM does not need to learn generalized representations, or learn-to-learn; instead, it needs to learn to perceive. This top-down, active perception interaction is compatible with theories of human cognition, where episodic recall is guided by semantic context \citep{greenberg2010interdependence}.

This type of interaction between a fast, episodic memory and a slow, statistical memory has also been explored in machine learning. In ``Human-inspired Episodic Memory for Infinite Context LLMs''  \citep{fountas2024emllm} augment a frozen large language model with an episodic memory module, retrieving past experiences to construct long-context reasoning without fine-tuning the model. 

Similarly, MemRL~\cite{zhang2026memrl} combines a frozen LLM with an episodic memory. Reinforcement signals are used to update the episodic memory and improve system performance without modifying the LLM. Both frameworks demonstrate that a fast memory system can leverage a slow, statistical model to interpret input in a way that supports rapid, context-dependent reasoning, providing a proof-of-concept for CLS-inspired active perception.

However, these works do not tackle continual, fast, few-shot learning, and do not construct a disentangled episodic STM for this purpose. Instead, they aim to improve retrieval augmented generation (RAG) task performance. In the cited works learning is still relatively slow and the resulting representations are still entangled.

%TODO - sparse recurrent encoding of sequences - stability without learning, combining information
%TODO - 
%system 1 and system 2 thinking
%learning about specific instances of classes
%can ensure stability by selectively attending to an actively perceived history in context. 

\section{Method}
To illustrate the ideas described above, we provide a small-scale implementation of a fast-learning, disentangled episodic STM as the primary actor in a complementary learning system. The memory is tested in streaming, zero-shot and few-shot settings to validate the concept that active perception and disentanglement allow continual learning without interference and forgetting of existing learning. For comparison, the memory will be compared to a conventional fully-connected feed-forward neural model.

%In this section, we describe the design and operation of the memory, explain how it interacts with the slow system, and provide a simple example illustrating its ability to rapidly store and recall relevant information in a controlled setting.

\subsection{Task}
The proposed system will be implemented as a Reinforcement Learning Agent which learns about a changing world in a series of episodes. Its only form of feedback will be environmental rewards. Model parameters must be updated after each environment step.

To implement the active-perception CLS architecture, the system will comprise an episodic STM which generates output actions and a slow, statistical, generalizing model as LTM which processes observations. 

The focus of our experiments is fast, zero and few-shot learning rather than replay and consolidation, which have already been demonstrated in many previous works such as \citep{kumaran2016learning,schapiro2017complementary,spens2024generative}. To explore STM few-shot learning capability, we used a frozen, pretrained Large Language Model (LLM) as the LTM. Since our experiments are relatively simple, it was sufficient to use a relatively small LLM. We used Mistral 7B~\citep{jiang2023mistral} via Ollama~\citep{ollama2023}. 

Source code for all methods including model and task generation is available at:

\url{https://github.com/drawlinson/disentangled_memory}

\subsubsection{Environment}
The test environment is presented as a stimulus-response challenge. Each episode is framed as an encounter with an object, plant or animal. The agent has up to 10 steps to decide how to respond to the encounter. During this time, it can issue queries to the LTM asking about the object. This framing was selected to make the agent's perceptions and actions relatable. In each step the agent can perform one of the following external actions instead of querying the LTM:

\begin{itemize}
    \item Do nothing (allows time for thinking before responding)
    \item Approach <object>
    \item Try to eat <object>
    \item Hide from <object>
    \item Run away from <object>
\end{itemize}

The STM is trained to model the expected returns of each action (including perceptual actions - see below). The agent as a whole is therefore a Q-learning model \citep{watkins1992qlearning}. An Epsilon-Greedy exploration strategy is used with $\epsilon=0.1$ during training.

Scenarios were generated by asking the LLM to generate a large set of entities from the prompt \textit{``You are a mouse in the garden and you see a ...''}. The 24 most common object classes were selected for inclusion in the dataset with frequency proportional to their occurrence in response to the prompt. They were grouped into four \textbf{broad} classes:

\begin{itemize}
    \item Maybe edible object
    \item A bird
    \item A land animal
    \item A plant
\end{itemize}

The broad class is available to the agent as an initial observation. Specific classes are not observable by the agent. They are:

\begin{multicols}{3}
    \begin{itemize}
        \item Cheese
        \item Tomato
        \item Deadly nightshade
        \item Carrot
        \item Slug pellets
        \item Cauliflower
        \item Fly agaric mushroom
        \item Radish
        \item Hawk
        \item Sparrow
        \item Eagle
        \item Pigeon
        \item Falcon
        \item Cat
        \item Dog
        \item Fox
        \item Snake
        \item Farmer
        \item Beetle
        \item Horse
        \item Mouse
        \item Capybara
        \item Tree
        \item Grass        
    \end{itemize}
\end{multicols}

For convenience, the dynamics of the environment are also generated by the same LLM. This is not part of its role as LTM but simply a way to generate an arbitrarily large set of scenarios for the Agent to reason about. Rewards are generated by asking the LLM questions about each encounter with <object> and what happens when specific actions are taken (see table \ref{tab:dynamics} for questions and resulting conditional rewards).

\begin{table}[htbp]
\caption{The agent receives a reward for its chosen action iff the LLM answers ``yes'' to the associated question. The LLM is queried for each instance of a scenario, allowing uncertainty in the answers. It is prompted with the scenario, the (unobservable) current encounter and then the specific question. So, for example, if <object> $=$ ``Fox'', the LLM is prompted: \textit{``You're a mouse in the garden and you see a Fox. Is it friendly? Answer yes or no.''} If the LLM answers with another token it is replaced with a ``?'' and no reward is applicable.}
\label{tab:dynamics}
\centering
\begin{tabular}{lllll}
    \toprule
   & \multicolumn{4}{l}{\textbf{Action}} \\
  \textbf{Question} & Approach <object> & Eat <object> & Hide & Run away \\
    \midrule
 Is it friendly? & 1 & & &\\
 Does it eat mice? & -1 & -1 & 1 & \\
 Is it edible? & & 1 & &\\
 Is it poisonous? & & -1 & &\\
 Does it chase mice? & & & & -1 \\
\bottomrule
\end{tabular}
\end{table}

\subsubsection{Withholding classes}
Novel encounters with unseen dynamics are modelled by withholding a set of specific classes from the training data. This allows analysis of zero-shot and few-shot learning of these withheld classes.

\subsubsection{Active perception interface}
In addition to the agent actions given above, the STM can also generate \textit{perception-actions} which are modelled as queries to the LTM. The LTM answers each query with a single token which is provided to the STM as input. The first token emitted by the LTM is always the broad class of the object encountered. It is assumed this is a readily observable quality of even an unseen object.

\begin{figure}[htbp]
     \centering
     % First Row
     \begin{subfigure}[b]{0.48\linewidth}
         \centering
         \includegraphics[width=\linewidth]{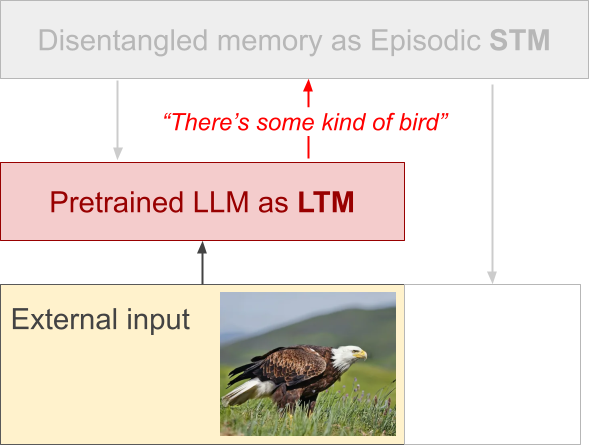}
         \caption{Initial observation is always the broad class of the object. The specific class of the object is not observable.}
         \label{fig:sub1}
     \end{subfigure}
     \hfill
     \begin{subfigure}[b]{0.48\linewidth}
         \centering
         \includegraphics[width=\linewidth]{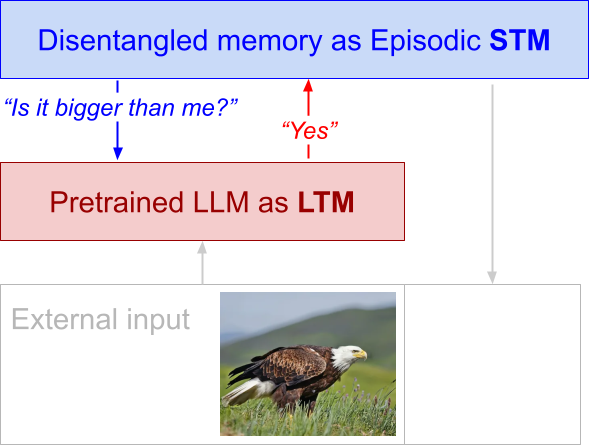}
         \caption{The STM can then use perception-actions to prompt the LTM to describe the object in general terms.}
         \label{fig:sub2}
     \end{subfigure}

     \vspace{1em} % Adds vertical space between rows

     % Second Row
     \begin{subfigure}{0.48\linewidth}
         \centering
         \includegraphics[width=\linewidth]{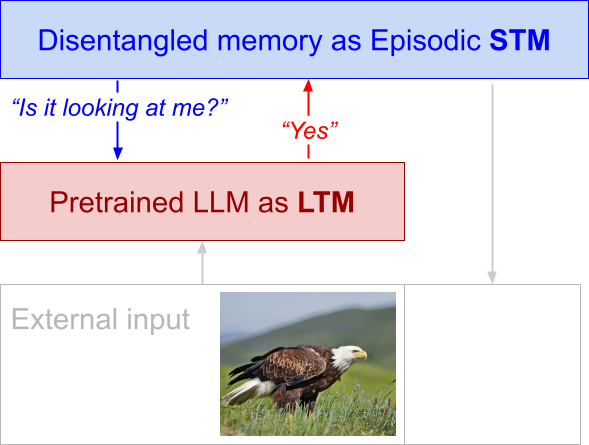}
         \caption{Responses from the LTM are accumulated in a context window and recurrently encoded into the STM input.}
         \label{fig:sub3}
     \end{subfigure}
     \hfill
     \begin{subfigure}[b]{0.48\linewidth}
         \centering
         \includegraphics[width=\linewidth]{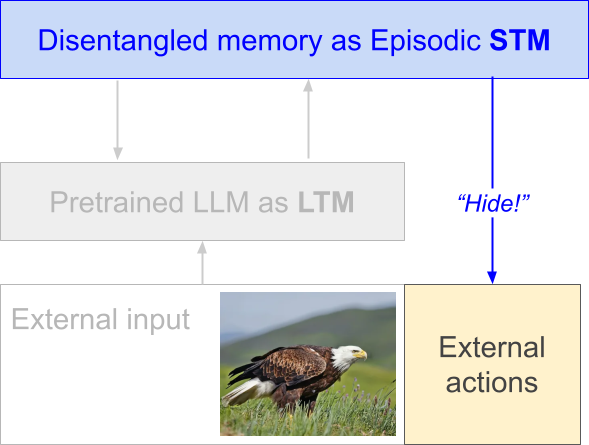}
         \caption{When ready, the STM can emit an external action, which may trigger an environmental reward (good or bad).}
         \label{fig:sub4}
     \end{subfigure}

     \caption{Each episode represents an encounter with an object. The agent must decide how to respond to the object, potentially resulting in a reward. The action space available to the STM includes a set of perceptual queries to the LTM, which will attempt to make generalized inferences about the stimulus.}
     \label{fig:active_perception_example}
\end{figure}

Over an episode, the STM receives a series of LTM tokens, allowing it to interrogate the LTM about different aspects of the encounter (see figure \ref{fig:active_perception_example}).

As in \citep{lampert2009learning}, the perception-actions available to the STM are all observable qualities of the object, which could be perceived even for novel objects ( the STM cannot ask the LTM for the specific class of the object):

\begin{multicols}{2}
    \begin{itemize}
        \item Does it look like a mouse?
        \item Is it bigger than me?
        \item Does it smell tasty?
        \item Does it have a long tail?
        \item Does it have four legs?
        \item Is it red?
        \item Is it green?
        \item Is it noisy?
        \item Is it watching me?
        \item Is it coming towards me?
    \end{itemize}
\end{multicols}

\subsection{Fast, episodic, Short-Term Memory}
This section describes our implementation of a fast learning STM with the desired qualities of disentangled and non-interfering, distributed representation. Disentanglement is achieved by using the active-perception action-space to select combinations of percepts - factors - from the LTM before processing and storing them in a recurrent, non-interfering sparse memory.

All variants of the STM can be trained in online, mini-batch or streaming fashion with no replay or buffering. The disentangled STM is demonstrated with a batch-size of 1 in some experiments. Models cannot use replay or back-propagation through time to assign rewards or losses to earlier states. This replicates the operating conditions of an embodied agent, but makes learning harder and slower (in terms of episodes).

The STM is required to estimate $Q(s,a)$ i.e. the expected return of actions $a$ in state $s$.

\subsubsection{Sparse recurrent encoder}
The STM receives a context window of the last $n=6$ LTM tokens as input. These tokens are transformed via a recurrent sparse encoder into a dense vector which is given as input to the STM.

Reasoning often requires an agent to integrate multiple observations over time. The number of possible observation \emph{sequences} grows exponentially with respect to context window length. This creates a problem for a fast-learning STM which cannot generalize.

Two features of active perception help to reduce the dimensionality of this context. Selective attention and filtering of irrelevant observations reduces the number of sequences experienced. This process is learned by the STM. In addition, perceptual generalization within the LTM can replace infinitely varied input with a finite number of labels, again reducing the number of potential contexts. Adding actions to reset the recurrent encoder and skip encoding of uninteresting input could also shrink the space of task-relevant recurrent encodings, but this feature is not used in this work.

This context must still be presented to the STM in a stable and deterministic manner to mitigate its lack of generalization ability. Recurrent sparse encoding of the input can preserve long-range dependencies, but these encodings generalize poorly \citep{gordon2020long, rawlinson2019learning}. Fortunately, in the proposed setting this doesn't matter, because generalization has already occurred during perception.

\subsubsection{Sparse Distributed Memory (SDM)}
An input consists of a dense real key vector $\bm{x}$ and a dense real value vector $\bm{v}$. In the Q-learning setting $\bm{x}$ is a concatenation of observation and action encodings and $\bm{v} = Q(s,a)$. To retrieve a predicted vector $\bm{\hat{v}}$ the key $\bm{x}$ is encoded with equation \ref{eqn:project}, which is simply a linear projection. Matrix $P$ has dimension $[m,n]$ where $m$ is the capacity of the memory and $n$ is the number of elements in the key vector $\bm{x}$. $M$ is initialized with random normal values scaled by  ${n}$.

\begin{equation}
    \bm{p} = P\bm{x}
    \label{eqn:project}
\end{equation}

To achieve high capacity, memory activation should be sparse and the values distributed. Equation \ref{eqn:topk} selects the $k$ indices $\bm{i}$  of $\bm{p}$ which have the largest values.

\begin{equation}
    \bm{i} = \operatorname{TopK}(\bm{p}, k)
    \label{eqn:topk}
\end{equation}

Since the target values are distributed among the top-k elements of the memory, equation \ref{eqn:sum} sums them to yield the predicted vector $\bm{\hat{v}}$. Elements of $V$ may participate in multiple active cliques for different inputs. To avoid instability during learning, equation \ref{eqn:delta} produces an error from the sum of all active elements' values, which allows contested elements' values to settle to an intermediate value or shrink to zero. Other elements' values adjust to reproduce the target vector $\bm{v}$ accurately. Representation of the target values is distributed among the active clique $\bm{i}$.

\begin{equation}
    \bm{\hat{v}} = \sum_{j=0}^{k} V_{\bm{i}_{j}}
    \label{eqn:sum}
\end{equation}

\begin{equation}
    \bm{\delta} = \frac{\bm{\hat{v}} - \bm{v}}{k} \eta
    \label{eqn:delta}
\end{equation}

\begin{equation}
    V \leftarrow V + \bm{\delta}
    \label{eqn:update}
\end{equation}

Equation \ref{eqn:update} uses $\bm{\delta}$ to update the stored values. Learning is therefore simply gradient descent. Since updates only affect values of the active clique, memories are highly orthogonal.

Only the values $V$ adapt during learning because projection $P$ is fixed. Due to high orthogonality and low interference, these updates may be arbitrarily fast without disrupting other stored values. Slower updates allow the memory to model the expected value of $V$ over multiple observations. Allowing this recency bias, the memory can learn in streaming, online mode or in minibatches.

To maximize memory capacity, small quantities can be randomly added to values of $P$ which correspond to memory ``cells'' that never participate in active cliques.

\subsubsection{Associative Sparse Distributed Memory}
SDM requires a complete key vector $\bm{x}$ to retrieve values. It is desirable to be able to retrieve and sample complete value vectors from incomplete cues (i.e. pattern completion). This capability can be achieved using a second instance of SDM which uses active cliques as its input. Together, the two memory models act as an associative, disentangled, sparse distributed memory. Although associated and distributed sound contradictory, both qualities exist simultaneously.

To make the system associative, the second memory stores active cliques. To ensure effective distributed representation, the input to the second memory must be a dense, entangled vector of real values. Equation \ref{eqn:sparse} converts the indices $\bm{i}$ of the active clique into a sparse vector $\bm{s}$, which is then projected via matrix $M$ into $\bm{d}$ which is the input for the second SDM. $M$ is initialized with random normal values scaled by $m$.

\begin{equation}
    \bm{s}_j = 
    \begin{cases} 
      1 & \text{if } j \in \bm{i} \\
      0 & \text{otherwise}
    \end{cases}
    \label{eqn:sparse}
\end{equation}

\begin{equation}
    \bm{d} = M\bm{s}
\end{equation}

Since $\bm{s}$ contains a 1 for each active member of observed cliques, complete cliques can be retrieved by cueing the second SDM with subsets. Meaningful values can then be retrieved from the original SDM using completed cliques. 

\subsection{Entangled, baseline STM model}
The disentangled STM was compared to a baseline model, namely a fully-connected 2-layer feed-forward network trained with the Adam optimizer. The model is expected to be entangled and degenerate during few-shot training due to this architecture, hence the name.

Like the disentangled STM, the entangled baseline model estimates the expected return $Q(s,a)$ for a given state, action pair. While the disentangled memory can learn with a batch size of 1, the entangled model requires a larger batch size for effective learning. Therefore, for most experiments the same batch size is used for both STM models.

We performed sequential univariate hyperparameter optimization on the entangled STM. Table \ref{tab:optimize} lists the parameter values which were empirically optimized. Optimized hyperparameters are described in Appendix \ref{app:param}.

\begin{table}[htbp]
\caption{Values explored during sequential univariate hyperparameter optimization to maximize entangled STM performance.}
\label{tab:optimize}
\centering
\begin{tabular}{ll}
    \toprule
  \textbf{Hyperparameter} & \textbf{Values evaluated} \\
    \midrule
 Learning rate & 0.01, 0.001, 0.0001, 0.00001 \\
 Input encoding & Concatenate and flatten, or recurrent encoding \\
 Hidden layer size & 500, 1000, 2000, 4000, 8000 \\
 Batch size & 8, 16, 32, 64 \\
 Dropout regularization & 0.1, 0.25, 0.3, 0.4 \\
 Input layer-norm & Enabled, disabled \\
\bottomrule
\end{tabular}
\end{table}

\section{Results}
\label{sec:results}
We report results using the same task and environment under different learning conditions: Few-shot learning, zero-shot learning, and streaming RL learning. All results are averaged over 8 independent runs except where otherwise stated.

\subsection{Few-shot results}
The purpose of this setting is to verify that the proposed disentangled memory can learn new data quickly without interference with existing memories (sometimes called ``catastrophic forgetting'') which typically occurs under these conditions when training entangled models \citep{mccloskey1989catastrophic, goodfellow2013empirical}. 

The setup for this experiment is a challenging form of few-shot learning \citep{antoniou2020defining} in which the model is first ``pre-trained'' on a set of data. Afterwards, the model is exposed to instances of new, unseen classes and allowed to learn from a few samples. The model is \textbf{never} allowed to train on the original data again. Instead, we evaluate performance on the new classes as the number of training exposures increases and continue to evaluate performance on the original data to verify that it does not decrease. In our experiment, ``forgetting'' would be evidenced by a reduction in mean reward per step.

The specific classes excluded from pre-training were ``deadly nightshade'' and ``fly agaric mushroom'' because these are both \emph{usually} reported by the LLM LTM to be red, foul-smelling plants which (unknown to the agent) are also poisonous. The training set contains some other plants which are red but not poisonous (e.g. ``tomato'' and ``radish''). Therefore, the optimal policy for red plants is initially to eat them, but the agent later learns only to eat certain red foods.

Performance in terms of mean-reward-per-step was measured after each additional set of exposures to the new data. Progressive exposure was measured at 10, 20, 40, 80, 160, 320, 640 and 1280 steps with batch size 16 (optimized for the entangled STM). Note that although few-shot exposures are counted in steps, it may take up to 10 steps for an episode to complete, \emph{possibly} yielding a reward; since the experiments are in an online setting without experience replay it may take many repeat episodes to associate delayed rewards with the perceptual actions that cause them.

\begin{figure}[htbp]
     \centering
     \includegraphics[width=\linewidth]{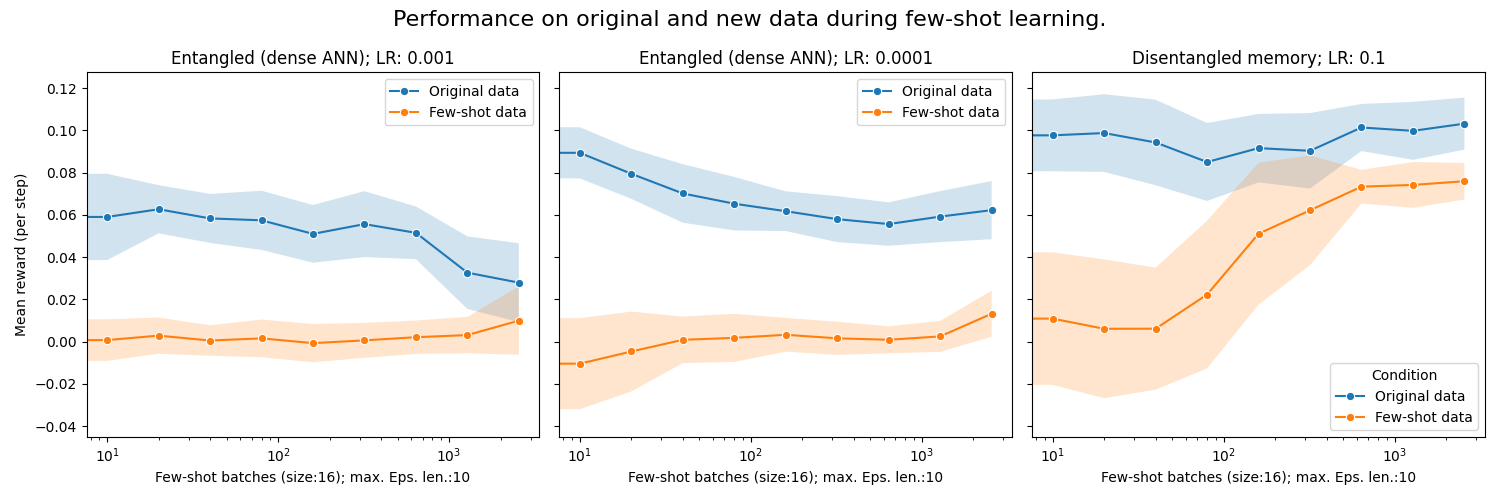}
     \caption{Effect of cumulative few-shot training exclusively on new data with no further exposure to the original data. ``Entangled'' models trained with stochastic gradient descent rapidly lose performance on the original data even before learning the new data. The proposed disentangled memory maintains performance (mean reward per step) on original data while rapidly learning the new data. Rewards only occur at the end of an episode, which may take up to 10 steps. Mean rewards of approximately 0.1 and 0.08 represent an optimal policy on the original and new data respectively.}
     \label{fig:few_shot_results}
\end{figure}

Figure \ref{fig:few_shot_results} shows that disentangled memory performance on the original data does not decrease even after thousands of training exposures to new data, remaining at approximately 0.1 which represents an optimal policy on this data.

\begin{figure}
  \begin{minipage}[c]{0.6\textwidth}
    \includegraphics[width=\textwidth]{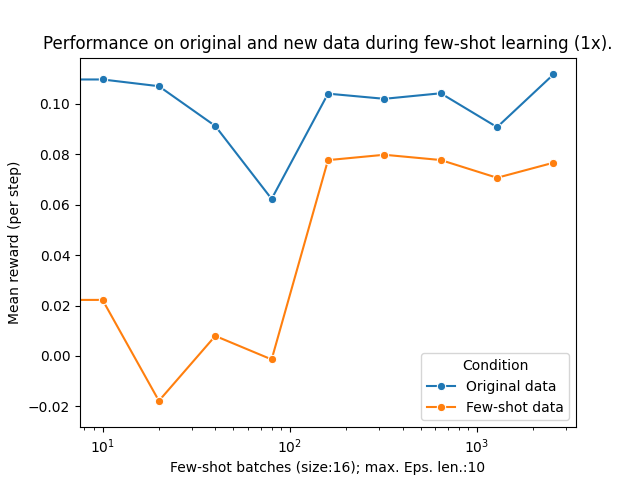}
  \end{minipage}\hfill
  \begin{minipage}[c]{0.4\textwidth}
    \caption{An interesting phenomenon during episodic few-shot learning, illustrated by the result of just one run. The model appears to forget the original data during learning of new encounters, producing a characteristic V-shaped drop in performance on the original data. However, this is actually caused by changes in policy during exploration of the new data. After discovery of the optimal policy for the new data, performance rebounds to its prior level on the original data, showing that nothing was forgotten after all.} 
\label{fig:few_shot_1x}
  \end{minipage}
\end{figure}

An interesting phenomenon observed in individual traces of disentangled few-shot learning is that the model appears to ``forget'' how to handle some encounters, and then seems to ``remember'' the optimal policy later (see figure \ref{fig:few_shot_1x}). In fact, the model has not forgotten anything, but as the optimal policy evolves, the \emph{value} of some actions varies. Once learning of new classes has completed, optimal policy is restored.

In contrast, entangled STM performance on the original data degrades from the same initial 0.1 mean reward to approximately 0.6 after 640 steps. This is the effect of the anticipated interference or ``forgetting''. Unlike the disentangled STM, entangled performance does not recover. In addition, the entangled STM is unable to learn as quickly, despite hyperparameter optimization including faster learning rates, which cause instability at 0.001. 

This baseline model was selected to demonstrate the anticipated effects of entanglement. Techniques such as Elastic Weight Consolidation \citep{Kirkpatrick2017EWC} can mitigate these effects, but do not eliminate them.

\subsubsection{Training time}
\begin{figure}
  \begin{minipage}[c]{0.6\textwidth}
    \includegraphics[width=\textwidth]{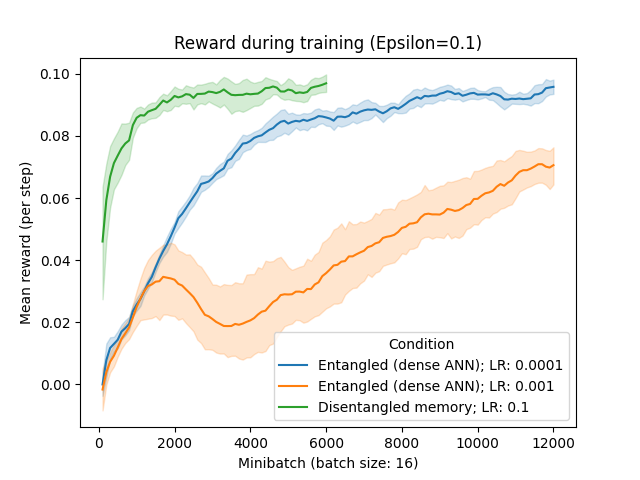}
  \end{minipage}\hfill
  \begin{minipage}[c]{0.4\textwidth}
    \caption{Performance in terms of mean-reward-per-step for each of the STM variants during training. The disentangled STM learns an optimal policy in approximately 2,000 minibatch steps. The entangled STM (a fully-connected, 2-layer feed-forward neural network) requires 12,000 steps to reliably reach the same performance. Attempting to increase learning rate to 0.001 causes instability and slower learning overall in the entangled STM.} 
\label{fig:training_results}
  \end{minipage}
\end{figure}

Figure \ref{fig:training_results} shows mean-reward-per-step during training on the original data. The disentangled STM reaches an optimal policy at around 2,000 steps, and training was discontinued at 6,000 steps. We did not optimize the learning rate parameter as it was already very fast; the limiting factor is exposure to a sufficient number of samples. 

The entangled STM learns more slowly, requiring 12,000 minibatch steps to reliably reach optimal policy with a learning-rate of 0.0001. Attempts to train the model more quickly with a learning-rate of 0.001 are also shown; paradoxically, this creates instability leading to slower learning overall.  

\subsection{Zero-shot results}
Zero-shot learning means to successfully perform a specific task without any specific experience of that task, typically by meta-learning \citep{finn2017model} or by leveraging experiences from other, related tasks \cite{lampert2009learning}.

The purpose of this experiment is to test whether the proposed mechanism of zero-shot learning in an active-perception CLS architecture actually works in practice. 

In theory, if the training data contains \emph{any} comparable encounters, the disentangled STM can zero-shot new encounters by generalizing from these experiences. Generalization occurs in the LTM which emits similar responses to various perceptual questions for both the training objects and the unseen zero-shot test objects. 

\begin{figure}[htbp]
     \centering
     % First Row
     \begin{subfigure}[b]{0.48\linewidth}
         \centering
         \includegraphics[width=\linewidth]{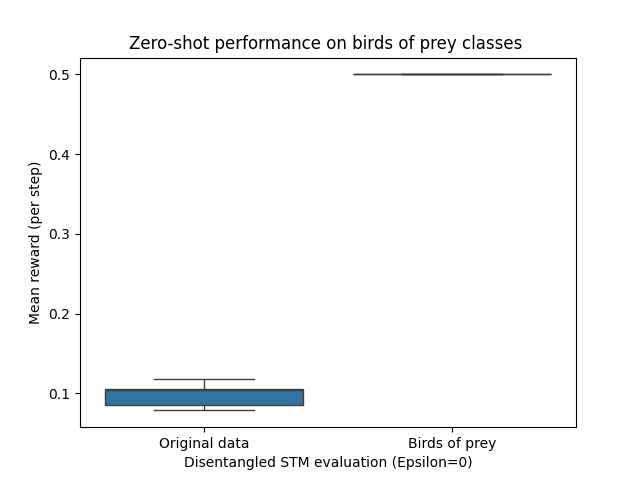}
         \caption{Zero-shot performance on unseen ``birds of prey'' encounters after pretraining on an original dataset which includes encounters with one type of predatory bird (mean reward 0.1). A mean reward of 0.5 per step represents optimal policy for bird-of-prey encounters. Although the LLM LTM which generates the percepts doesn't give identical answers for Falcons, Eagles and Hawks, individual answers (factors) are consistent enough that the STM knows exactly what to do, despite being unable to generalize.}
         \label{fig:zero_shot_results}
     \end{subfigure}
     \hfill
     \begin{subfigure}[b]{0.48\linewidth}
         \centering
         \includegraphics[width=\linewidth]{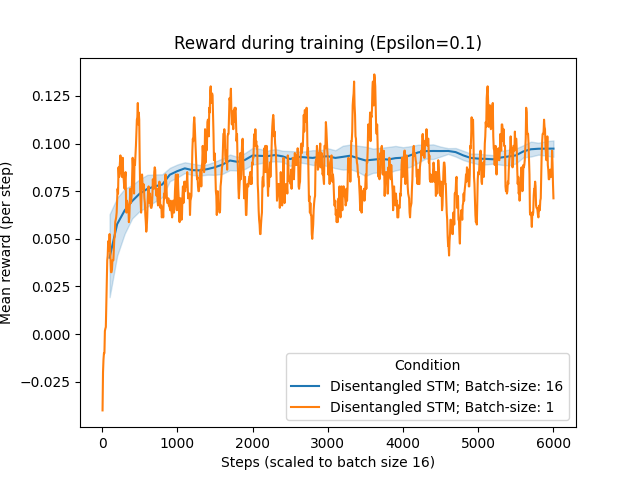}
         \caption{Comparison of model performance between training on batch-size 16 and one training run with batch-size 1 (streaming RL). Given comparable exposures, mean reward is similar in both conditions (the number of steps in the streaming condition is divided by 16). This result demonstrates that the model can learn quickly from a single thread of experience. This is relevant to applications such as robotics, where it would allow a robot to rapidly adapt to specific situations.}
         \label{fig:online_results}
     \end{subfigure}
     
     \caption{Disentangled STM performance under zero-shot and streaming conditions.}  
     \label{fig:zero_shot_online_results}
\end{figure}

For example, figure \ref{fig:zero_shot_results} compares disentangled STM performance on bird-of-prey encounters (``hawk'', ``eagle'', ``falcon'') when trained on all object instances and when ``eagle'' and ``falcon'' are withheld from the training set. The LLM LTM does not give identical answers to all the perceptual questions about these birds; for example, it often claims that eagles have long tails and falcons don't. However, the disentangled STM has learned these features are irrelevant to the danger.

A mean reward of 0.5 per step represents optimal policy; in fact the model learns to hide from all birds. Note that performance is unaffected by the absence of the latter classes from the training set, verifying the concept of zero-shot learning via perceptual generalization is feasible.

\subsection{Streaming RL results}
An online Reinforcement Learning (RL) setting is one where the agent interacts with an environment (possibly simulated) rather than a finite, pre-recorded dataset\citep{sutton1998reinforcement}. However, most online RL algorithms also make extensive use of replay buffers for more sample-efficient learning and less biased sampling \citep{mnih2015human}.

Streaming RL is a more restrictive subset of online learning where data must be processed immediately and then discarded. Therefore, policy must also be updated after every step. It mimics natural intelligence by learning continuously without a large experience replay buffer. Recently, careful analysis of optimizer behaviour, layer normalization, weight initialization and the use of eligibility traces has led to significantly improved model performance in streaming settings~\cite{elsayed2024streaming}.

This experiment is simply a demonstration that the disentangled STM produces comparable performance with a mini-batch size of 1 or 16 (the latter value used in few-shot experiments for the benefit of the entangled STM). In conjunction with the other requirements of policy update every step and no replay buffer, this fits the criteria for streaming RL. Figure \ref{fig:online_results} shows that mean reward per batch after equivalent training exposures is indeed comparable regardless of batch size. 

This capability potentially allows a fast learning STM to learn from a singular autobiographical experience, and to learn to reason about specific instances of objects or situations \citep{kowadlo2024expanding}, a topic which has not received much attention in the literature. However, it could be very useful in robotics where it might allow embodied agents to rapidly adapt to specific, individual circumstances. 

\section{Conclusion}
The experiments described in this paper present a minimally complex yet relatable proof-of-concept for the core motivating idea that abandoning generalization allows an episodic STM to learn quickly and continuously, and that active perception can allow the agent to overcome a lack of generalization ability in some components. The experiments place an autonomous agent in a series of encounters and measure it's ability to rapidly learn new responses to unseen object classes without interfering with existing memories.

Many ideas work in theory, but cannot be realized. While this demonstration does not attempt to benchmark performance on large-scale tasks or compare against state-of-the-art continual learning systems, it does provide evidence that the core concepts of disentanglement, generalization via active perception and non-interference through sparse distributed encoding collectively enable the desired qualities of continual, streaming, zero and few-shot learning. Further work is needed to discover how to scale up these techniques to more complex and realistic high-dimensional environments. 

Our intent is to apply this architecture to the ARC-AGI-3 challenge, which as a few-shot, interactive game environment without supervised feedback is surprisingly compatible with the settings described in this paper \citep{arcprize2026arcagi3}. The ARC-AGI challenge evolved from Chollet's original ARC corpus, which was proposed specifically to challenge some of the weaknesses prevalent in modern machine learning, such as slow-learning and assumed i.i.d. sampling of large, stationary datasets \citep{chollet2019measure}.

\subsection{Future work}
In addition to scaling, the proposed non-interfering, fast-learning memory can be used in other RL architectures. The associative SDM supports pattern completion, which allows sampling from the memory. Although not demonstrated in this paper, this variant can be used to build fast-learning world-models which rapidly adapt to specific circumstances. These world models can be exploited to further increase sample efficiency, because rollouts in these world-models allow policy or value model updates without repeated interaction with the environment - potentially another mechanism for zero or few-shot learning.

The perceptual action-space used in this paper was specified in advance and was not learned from the data. In a more sophisticated version of this architecture, the STM should automatically discover useful perceptual frames and queries in the latent space of the LTM.

Finally, although replay and consolidation into LTM have previously been extensively investigated, it would be desirable to demonstrate specifically that the disentangled STM can generate suitable sequences for replay and consolidation.

%%%%%%%%%%%%%%%%%%%%%%%%%%%%%%%%%%%%%%%%%%%%%%%%%%%%
%%%%%%%%%%%%%%%%%%%%%%%%%%%%%%%%%%%%%%%%%%%%%%%%%%%%

%Of note is the command \verb+\citet+, which produces citations appropriate for use in inline text.  For example,
%\begin{verbatim}
%   \citet{hasselmo} investigated\dots
%\end{verbatim}

\begin{ack}
Thanks to Alan Zhang and Long Dang for inspiring conversations and tolerating months of David's complaints about what's wrong with AI, which led directly to this approach to fast, continual learning.
\end{ack}

\bibliographystyle{abbrvnat}
\bibliography{bibliography}

%%%%%%%%%%%%%%%%%%%%%%%%%%%%%%%%%%%%%%%%%%%%%%%%%%%%%%%%%%%%
\clearpage
\appendix

\section{Appendix / supplemental material}
\label{app:param}
Tables \ref{tab:common}, \ref{tab:disentangled} and \ref{tab:entangled} contain hyperparameters used for each STM variant.

\begin{table}[H]
  \caption{Common hyperparameters}
  \label{tab:common}
  \centering
  \begin{tabular}{ll}
    \toprule
    Name     & Value      \\
    \midrule
    Mini-batch size & 16 \\
    Evaluation steps & 1000 \\
    Observation context window & 6 \\
    Maximum episode length (steps) & 10 \\
    Token embedding size & 100 \\
    Recurrent observation embedding size & 200 \\
    Action embedding size & 200 \\
    Discount factor & 0.9 \\
    $\epsilon$ & 0.1 \\
    \bottomrule
  \end{tabular}
\end{table}

\begin{table}[H]
  \caption{Disentangled STM hyperparameters}
  \label{tab:disentangled}
  \centering
  \begin{tabular}{ll}
    \toprule
    Name     & Value      \\
    \midrule
    Observation encoding & Recurrent \\
    Memory capacity $m$ & 10000 \\
    Activation sparsity $k$ & 32 \\
    Learning rate & 0.1 \\
    \bottomrule
  \end{tabular}
\end{table}

\begin{table}[H]
  \caption{Entangled STM hyperparameters}
  \label{tab:entangled}
  \centering
  \begin{tabular}{ll}
    \toprule
    Name     & Value      \\
    \midrule
    Observation encoding & Flatten \\
    Learning rate & 0.0001 \\
    Layers & 2 \\
    Bias & Enabled \\
    Hidden layer size & 2000 \\
    Hidden nonlinearity & Leaky-ReLU \\
    Input Layer-norm & True \\
    Dropout rate & 0.25 \\
    Weight clipping & Disabled \\
    \bottomrule
  \end{tabular}
\end{table}

%%%%%%%%%%%%%%%%%%%%%%%%%%%%%%%%%%%%%%%%%%%%%%%%%%%%%%%%%%%%

\end{document}